\pdfoutput=1

\documentclass[11pt]{article}

\usepackage[final]{acl}

\usepackage{times}
\usepackage{latexsym}
\usepackage{booktabs}

\usepackage[T1]{fontenc}

\usepackage[utf8]{inputenc}

\usepackage{microtype}

\usepackage{inconsolata}

\usepackage{graphicx}

\usepackage{listings}
\usepackage{caption}
\usepackage{geometry}
\title{PLUGH: A Benchmark for Spatial Understanding and Reasoning in Large Language Models}

\author{Tikhonov Alexey \\
  Inworld.AI, Germany \\
  \texttt{altsoph@gmail.com}
}

\begin{document}
\maketitle
\begin{abstract}
We present PLUGH\footnote{\url{https://www.urbandictionary.com/define.php?term=plugh}}, a modern benchmark that currently consists of 5 tasks, each with 125 input texts extracted from 48 different games and representing 61 different (non-isomorphic) spatial graphs to assess the abilities of Large Language Models (LLMs) for spatial understanding and reasoning. Our evaluation of API-based and open-sourced LLMs shows that while some commercial LLMs exhibit strong reasoning abilities, open-sourced competitors can demonstrate almost the same level of quality; however, all models still have significant room for improvement. We identify typical reasons for LLM failures and discuss possible ways to deal with them. Datasets and evaluation code are released\footnote{\url{https://github.com/altsoph/PLUGH}}.
\end{abstract}

\section{Introduction}
Large Language Models (LLMs) have shown remarkable capabilities, but there are still tasks that they struggle with. Spatial reasoning, understanding, and planning are among these challenging tasks \cite{2305.15771, 2309.15577, CogEval}. In modern projects, this functionality is often implemented using external mechanisms such as Cognitive Architectures \cite{2309.02427}. For example, in Simulacra \cite{Simulacra}, a spatial tree is explicitly queried, and prompts are formed based on the results.

As the quality of LLMs improves, evaluating their progress becomes increasingly difficult \cite{tikhonov2023post}. However, we expect improvements in these aspects. To track this growth, we propose using our benchmark, which is aimed at assessing the quality of understanding the structure of space described in arbitrary fictional texts. 
Reconstructing such structures from text is a complex task \cite{Ammanabrolu_Cheung_Tu_Broniec_Riedl_2020, ammanabrolu2021learning}. On the one hand, spatial language understanding involves recognizing and reasoning about spatial semantics, e.g., spatial objects, relations, and transformations, in natural language descriptions \cite{Battista}. On the other hand, extracting formal knowledge from fictional text implies narrative understanding \cite{2310.18783}.

The challenge, however, is how to objectively create ground truth annotations, as the concept of location and transitions between locations is inherently abstract and context-dependent.

Existing works \cite{2309.15577, SPARTQA, CogEval, Floorplan} usually solve this problem by generating formal descriptions from graphs, but such descriptions are often monotonous and lack variability.  

Recently, text-based games have been frequently used as a unique source of information or as agents' playgrounds \cite{Jericho, TextWorld, LIGHT}, as they bridge the gap between formally functioning text-based interactive environments and fictional texts. Text-based games are also often used for generating benchmarks \cite{pan2023rewards, tan2023textbased, tikhonov-2024-branching}.
In \cite{Peng_Cui_Zhou_Jia_Riedl_2023}, agents are trained to extract Knowledge Graphs from text stories, and they also use text-based games for data generation.

We propose using text-based games to construct such a benchmark, specifically a collection of games with known walkthroughs suitable for running on the Jericho emulator \cite{Jericho}, resulting in pairs of diverse fictional texts and formal spatial structures.

\begin{figure*}[th]
  \includegraphics[width=\textwidth]{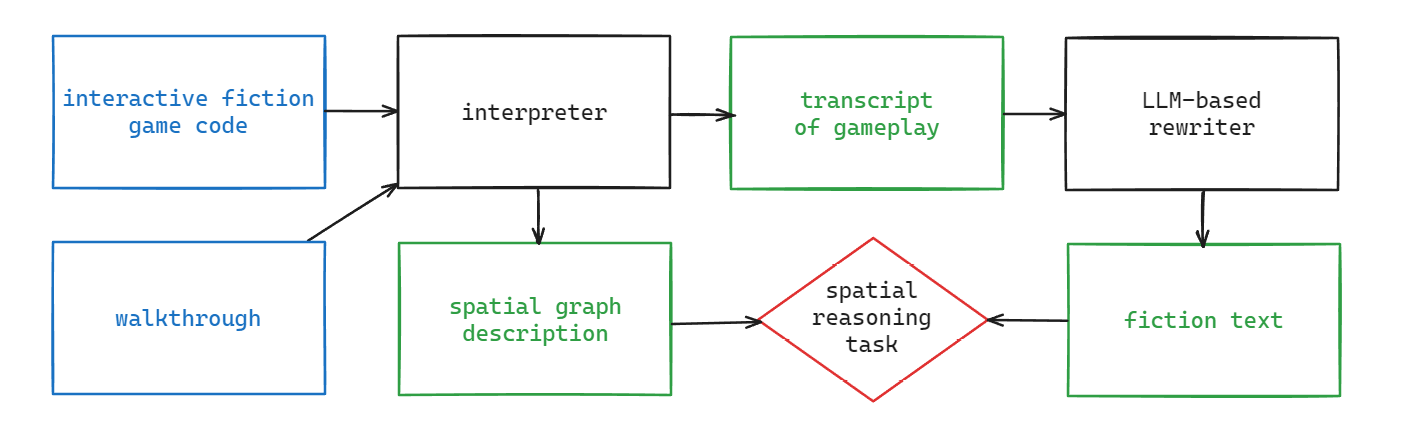}
  \caption{The principal schema of our approach.}
  \label{fig:schema}
\end{figure*}

This work contributes by:
\begin{itemize}
    \item Proposing a formal spatial reconstruction and reasoning benchmark with 5 different tasks:
    \begin{itemize}
        \item Task 1: Graph reconstruction
        \item Task 2a: Character's path reconstruction
        \item Task 2b: Reversed character's path reconstruction
        \item Task 3: Novel shortest path
        \item Task 4: Temporal hinted shortest path
    \end{itemize}
    \item Introducing the PLUGH dataset with data crafted to facilitate these tasks.
    \item Demonstrating the effectiveness of modern models in solving these tasks and analyzing typical reasons for failures.
\end{itemize}

\section{Approach}

Text-based games represent a unique source of such information:
\begin{itemize}
    \item On the one hand, the player interacts with a partially observable, modeled environment through actions and observations conveyed in natural language. Thus, the game transcript is quite close to a natural linear fictional text (and can be transformed into one using modern LLMs).
    \item On the other hand, the game code strictly and formally defines the list of available locations and transitions between them. This information is already described by the game authors and can be extracted and used as ground truth.
\end{itemize}

\begin{figure}[t]
  \includegraphics[width=\columnwidth]{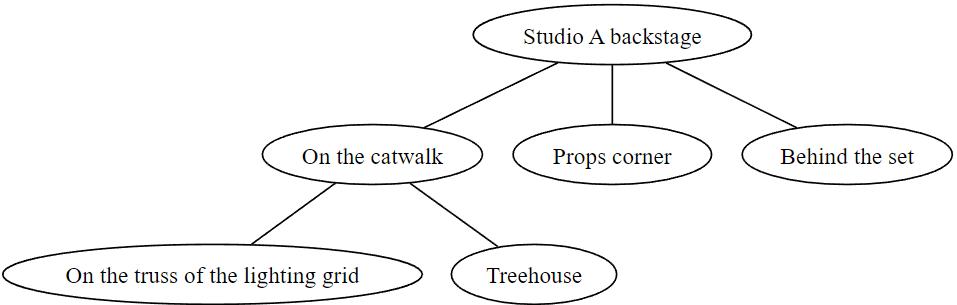}
  \caption{An example of a good segment from the Asgard game passed all filters.}
  \label{fig:good-segment-map}
\end{figure}

The general schema of our approach is presented in Figure \ref{fig:schema}.
\begin{itemize}
    \item We used the Jericho engine \cite{Jericho} and several dozen available games with known walkthroughs. 
    \item By replaying the walkthrough, we simultaneously obtained the game transcript and the spatial graph (checking for location changes after each command). 
    \item Since full-size transcripts and graphs turned out to be very diverse in size (from 3 to several hundred locations), we used a sliding window logic to find "good" segments of the walkthrough.
    \item A segment is considered good if its corresponding graph has 6 to 20 nodes, is connected, and sufficiently non-degenerate: the total number of basic cycles and leaf nodes should be at least 4 (thus excluding trivial linear graphs). An example of a good segment is shown in Figure \ref{fig:good-segment-map}.
    \item Then, the transcript of each segment was rewritten using the GPT-4 model into a fiction text (see the used prompt in Appendix \ref{sec:appendixA}). An example of the text before and after rewriting is shown in Appendix \ref{sec:appendixB}.
\end{itemize}


To validate the graph-text pairs, we performed the following checks:
\begin{itemize}
    \item The text should contain all node names as substrings.
    \item The graph should not contain duplicate nodes or nodes whose names are substrings of each other.
\end{itemize}

As a result of this filtering, we obtained 125 segments from 48 unique games. The average graph size is 8.64 nodes (see more statistics in Table \ref{tab:tab0})
resulting in 61 mutually non-isomorphic graphs (excluding node labels); see Appendix \ref{sec:appendixD}.
\begin{table}[t]
\centering
\small
\begin{tabular}{|cc|cc|cc|} 
\hline
\textbf{nodes} & \textbf{\#} & \textbf{edges} & \textbf{\#} & \textbf{cycles} & \textbf{\#} \\
\hline
6    & 39 & 5    & 38 & 0    & 94 \\
7    & 12 & 6    & 10 & 1    & 17 \\
8    & 14 & 7    & 16 & 2    & 2  \\
9    & 20 & 8    & 17 &      &    \\
10+  & 28 & 9+   & 32 &      &    \\
\hline
\end{tabular}
\caption{Overview of graph population statistics}
\label{tab:tab0}
\end{table}
Twelve segments (marked as a "dev" split) were used to generate few-shot examples for the remaining segments.

\section{Tasks and Results}
In this section we describe 5 novel tasks based on the data we collected and provide results of modern LLMs evaluation, including 3 models from OpenAI \cite{openai2024gpt4}, the latest model from Anthropic \cite{claude3} and open-source models, LLaMa3 \cite{llama3modelcard} and Mixtral \cite{jiang2024mixtral}.

\subsection{Task 1: Graph Reconstruction}
\textbf{Task Description:} \textit{You will be provided with a short fiction text. Your task is to extract the mentioned locations and compile a description of the locations graph in a graphviz format, undirected, without node descriptions, only with edges without labels for directly connected nodes.}

\textbf{Target Data:} Ground truth graph description.

\textbf{Metrics:} F1 scores for nodes and edges retrieval, \textit{the higher the better}.
\\

\textbf{Notes:}\begin{itemize}
    \item Some models ignore some instructions, providing, for example, directed graphs or graphs with edge labels. Adding a few examples to the prompt usually significantly improves it.
    \item It's not always possible to get identical node names (sometimes there are several options to name one node). To minimize the effects of these issues, we provided a relaxed ("fuzzy") parsing and matching algorithm. However, we ensure there are no false positives in that matching by avoiding similar location names in ground truth data and requiring that one of the matching items be a substring of another. 
    \item One may want to use different graph reconstruction metrics (for example, see Table \ref{table:performance_comparison}), so we provide a modular code that can be used to add any additional metrics.
\end{itemize}

Results for Task 1 for several modern models are presented in Table \ref{tab:tab1}. We find no clear dependency pattern between task 1 metrics and graphs' properties (like nodes, edges, and cycles) on available data.

\begin{table}[ht]
\small
\centering
\begin{tabular}{lcccc}
\toprule
\textbf{Model} & \textbf{0-shot} & \textbf{1-shot} & \textbf{2-shot} & \textbf{3-shot} \\
\midrule
gpt-3.5-turbo & 41.4\% & 54.1\% & 60.4\% & 62.2\% \\
gpt-4-turbo & \textbf{66.8\%} & 70.7\% & 70.8\% & 71.4\% \\
gpt-4o & \textbf{66.8\%} & 67.7\% & 70.7\% & 72.0\% \\
claude-3-opus & 64.2\% & \textbf{73.9\%} & \textbf{76.0\%} & \textbf{78.8\%} \\
llama3-8b & 55.4\% & 50.6\% & 53.7\% & 56.8\% \\
mixtral-8x7b & 17.0\% & 53.0\% & 55.3\% & 58.3\% \\
llama3-70b & 61.2\% & 67.9\% & 69.6\% & 70.0\% \\
mixtral-8x22b & 59.9\% & 68.0\% & 70.4\% & 68.5\% \\
\bottomrule
\end{tabular}
\caption{Graph Reconstruction task: edges retrieval F1-score, the higher the better.}
\label{tab:tab1}
\end{table}

\subsection{Task 2a: Character's Path Reconstruction}
\textbf{Task Description:} \textit{You will be provided with a short fiction text and a list of location names. Your task is to extract the main character's path as a sequence of visited locations, one by one, each on a new line.}

\begin{figure}[t]
  \includegraphics[width=\columnwidth]{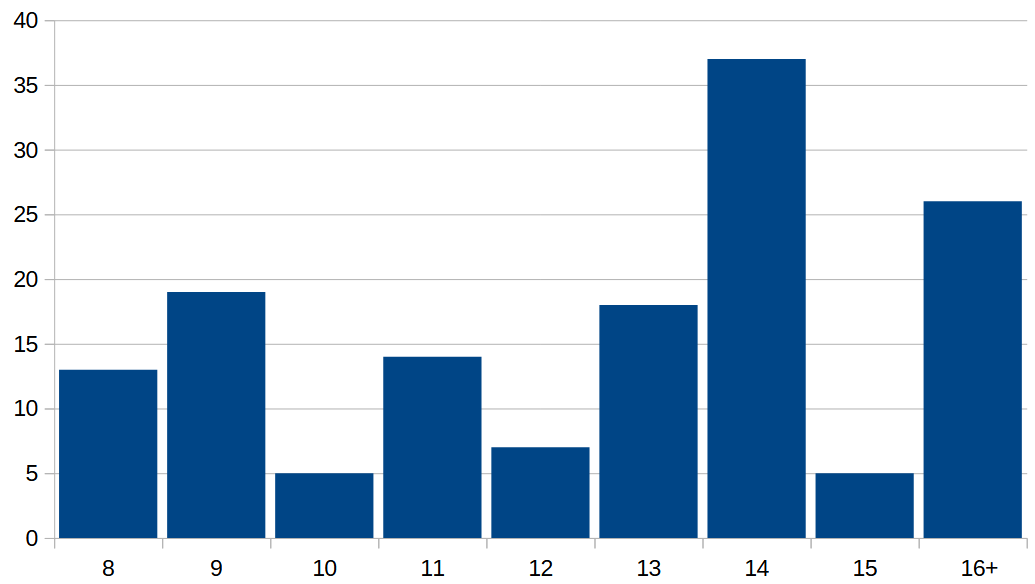}
  \caption{The distribution of path
lengths across the graphs in tasks 2a and 2b.}
  \label{fig:t2d}
\end{figure}

\textbf{Target Data:} Ground truth path (according to the locations sequence in the walkthrough). The distribution of path lengths across the graphs in Task 2 is provided in Fig \ref{fig:t2d}. 

\textbf{Metric:} Normalized Levenshtein distance, \textit{the lower the better}.

Results for Task 2a for several modern models are presented in Table \ref{tab:task2a}.

\begin{table}[ht]
\small
\centering
\begin{tabular}{lcccc}
\toprule
\textbf{Model} & \textbf{0-shot} & \textbf{1-shot} & \textbf{2-shot} & \textbf{3-shot} \\
\midrule
gpt-3-5-turbo & 33.6\% & 30.5\% & 28.5\% & 28.2\% \\
gpt-4-turbo & 16.2\% & 12.8\% & 11.9\% & 11.7\% \\
gpt-4o & \textbf{12.7\%} & \textbf{9.5\%} & \textbf{9.3\%} & \textbf{9.3\%} \\
claude-3-opus & 19.2\% & 11.1\% & 10.5\% & 10.0\% \\
llama3-8b & 38.7\% & 51.3\% & 44.1\% & 46.4\% \\
mixtral-8x7b & 35.9\% & 36.6\% & 38.2\% & 40.3\% \\
llama3-70b & 23.4\% & 14.9\% & 14.4\% & 15.8\% \\
mixtral-8x22b & 18.7\% & 17.7\% & 16.4\% & 20.5\% \\
\bottomrule
\end{tabular}
\caption{Character's Path Reconstruction task: normalized Levenshtein distance, the lower the better.}
\label{tab:task2a}
\end{table}

\subsection{Task 2b: Reversed Character's Path Reconstruction}
\textbf{Task Description:} Same as Task 2a, but we ask models to produce the reversed path.

Results for Task 2b for several modern models are presented in Table \ref{tab:task2b}.

\begin{table}[ht]
\small
\centering
\begin{tabular}{lcccc}
\toprule
\textbf{Model} & \textbf{0-shot} & \textbf{1-shot} & \textbf{2-shot} & \textbf{3-shot} \\
\midrule
gpt-3.5-turbo & 64.5\% & 60.2\% & 58.4\% & 59.5\% \\
gpt-4-turbo & 21.7\% & 18.2\% & 15.2\% & 15.5\% \\
gpt-4o & \textbf{19.4\%} & \textbf{13.0\%} & \textbf{11.7\%} & \textbf{12.4\%} \\
claude-3-opus & 25.7\% & 23.6\% & 21.1\% & 21.0\% \\
llama3-8b & 65.0\% & 63.6\% & 62.2\% & 61.7\% \\
mixtral-8x7b & 60.8\% & 54.4\% & 57.2\% & 55.5\% \\
llama3-70b & 37.2\% & 32.1\% & 30.2\% & 32.8\% \\
mixtral-8x22b & 46.1\% & 40.8\% & 39.8\% & 41.5\% \\
\bottomrule
\end{tabular}
\caption{Reversed Character's Path Reconstruction task: normalized Levenshtein distance, the lower the better.}
\label{tab:task2b}
\end{table}

\subsection{Task 3: Novel Shortest Path}
\textbf{Task Description:} \textit{You will be provided with a short fiction text and a list of location names. Your task is to extract the shortest path between two given locations (source and target) as a sequence of visited locations starting from the source and ending with the target location, one by one, each on a new line}.

\textbf{Target Data:} Shortest path between source and target locations (list of them if there are several shortest paths). In all cases, the length of the target path was 3 nodes.

\textbf{Metric:} Normalized Levenshtein distance, \textit{the lower the better}.

\textbf{Notes:} We selected the pair of locations with the following properties:
\begin{itemize}
    \item There is at least one path from one to another in the segment's graph.
    \item Each path has at least 3 steps (so we do not use directly connected locations).
    \item At least one of these paths wasn't presented in the transcript.
\end{itemize}

Results for Task 3 for several modern models are presented in Table \ref{tab:task3}.

\begin{table}[ht]
\small
\centering
\begin{tabular}{lcccc}
\toprule
\textbf{Model} & \textbf{0-shot} & \textbf{1-shot} & \textbf{2-shot} & \textbf{3-shot} \\
\midrule
gpt-3.5-turbo & 30.4\% & 21.6\% & 21.2\% & 20.6\% \\
gpt-4-turbo & \textbf{18.6\%} & \textbf{13.1\%} & \textbf{12.4\%} & \textbf{11.8\%} \\
gpt-4o & 26.2\% & 15.9\% & 13.2\% & 13.0\% \\
claude-3-opus & 37.7\% & 23.0\% & 20.2\% & 17.4\% \\
llama3-8b & 53.4\% & 43.3\% & 51.9\% & 69.7\% \\
mixtral-8x7b & 37.4\% & 34.1\% & 21.5\% & 20.4\% \\
llama3-70b & 38.3\% & 15.3\% & 16.2\% & 14.1\% \\
mixtral-8x22b & 33.2\% & 17.2\% & 15.9\% & 15.8\% \\
\bottomrule
\end{tabular}
\caption{Novel Shortest Path task: normalized Levenshtein distance, the lower the better.}
\label{tab:task3}
\end{table}

\subsection{Task 4: Temporal Hinted Shortest Path}
\textbf{Task Description:} \textit{You will be provided with a short fiction text and a list of location names. Your task is to extract the shortest path between two given locations (source and target) as a sequence of visited locations starting from the source and ending with the target location, one by one, each on a new line}.

\begin{figure}[t]
  \includegraphics[width=\columnwidth]{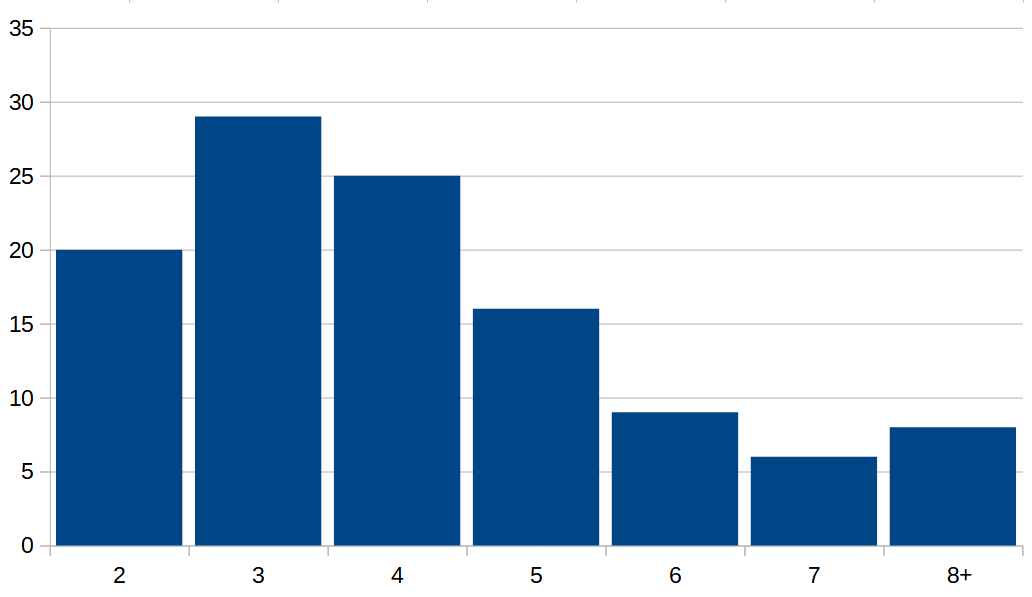}
  \caption{The distribution of path
lengths across the graphs in Task 4.}
  \label{fig:t4d}
\end{figure}

\textbf{Target Data:} Shortest path between source and target locations (list of them if there are several shortest paths). The distribution of path lengths across the graphs in Task 4 is provided in Fig \ref{fig:t4d}. 

\textbf{Metric:} Normalized Levenshtein distance, \textit{the lower the better}.

\textbf{Notes:} We selected the pair of locations with the following properties:
\begin{itemize}
    \item There is at least one path from one to another in the segment's graph.
    \item Each path has at least 3 steps (so we do not use directly connected locations).
    \item At least one of these paths wasn't presented in the transcript.
    \item Instead of explicitly specifying the starting and ending locations, we use hints about where an object was first encountered or last seen in the narrative. For example, \textit{"the place of the first encounter of nugget"} or \textit{"the place where cereal was left."}
\end{itemize}

Results for Task 4 for several modern models are presented in Table \ref{tab:task4}.

\begin{table}[ht]
\small
\centering
\begin{tabular}{lcccc}
\toprule
\textbf{Model} & \textbf{0-shot} & \textbf{1-shot} & \textbf{2-shot} & \textbf{3-shot} \\
\midrule
gpt-3.5-turbo & 66.6\% & 60.3\% & 51.9\% & 53.1\% \\
gpt-4-turbo & 30.2\% & 17.9\% & 20.0\% & 20.7\% \\
gpt-4o & \textbf{24.4\%} & \textbf{15.9\%} & \textbf{14.8\%} & \textbf{14.7\%} \\
claude-3-opus & 38.7\% & 30.2\% & 25.7\% & 21.6\% \\
llama3-8b & 70.6\% & 67.8\% & 79.7\% & 87.2\% \\
mixtral-8x7b & 64.5\% & 62.8\% & 60.9\% & 60.9\% \\
llama3-70b & 44.8\% & 31.4\% & 30.4\% & 31.0\% \\
mixtral-8x22b & 46.3\% & 34.8\% & 32.6\% & 31.4\% \\
\bottomrule
\end{tabular}
\caption{Temporal Hinted Shortest Path task: normalized Levenshtein distance, the lower the better.}
\label{tab:task4}
\end{table}

\section{Evaluation}

The evaluation results provided in the previous section were calculated using generations sampled with temperature 0.01 (since some of the models deny usage of zero temperature). Before comparison to the target data we prepocess and normalize sampled responses, to deal with some technical issues, like incorrect graph specification format or ambiguity in location naming (see the next subsection for more details).

Although different tasks rank models differently, one can speculate on the reasons and underlying patterns:
\begin{itemize}
    \item In Task 1, which can be considered a task of spatial summarization, the leader was Claude-3-opus, followed closely by GPT-4 and GPT-4o with minor differences.
    \item In the other tasks, which can be considered tasks of spatial reasoning, the leaders on average were GPT-4 and GPT-4o, with Claude-3-opus slightly behind.
    \item In Task 2b and Task 4, which requires additional reasoning (reversing the path and tracking object locations, correspondingly), all models except GPT-4o and GPT-4 showed significant degradation.
    \item Among tested open-source models, LLaMA3 70B led, with Mixtral 8x22B following closely.
    \item Smaller models (including GPT-3.5) found it significantly more challenging to understand the task setup, so their quality was substantially lower in the 0-shot setting. Still, it improved noticeably with few-shot examples.
    \item Even the top models are still imperfect in solving any of the presented tasks, leaving room for further improvement. The following subsection analyzes some typical errors and discusses their origins, pointing out possible ways for future advancements.
\end{itemize}
\subsection{Analysis of Typical Errors}
In this subsection, we analyze typical problems that arise in models when solving the problems proposed in this paper and discuss how such issues can be detected and addressed in practice.
\begin{itemize}
    \item \textbf{Incorrect formatting:} Sometimes models, especially smaller ones, tend to ignore part of instructions, generating, in particular, directed graphs or graphs with extra information (e.g., labels on edges) (see Table \ref{table:graphviz_comparison} for example). To overcome this problem, we introduced a flexible parsing code that we used to preprocess generated texts (see implementation code).

\begin{table}[t]
    \centering
    \begin{tabular}{|c|}
        \hline
        \textbf{Correct} \\
        \hline
        \begin{lstlisting}[basicstyle=\ttfamily\tiny]
graph G {
    "y2" -- "low n/s passage";
    "dusty rock room" -- "complex junction";
    "west end of twopit room" -- "east end of twopit room";
    "plover room" -- "y2";
    "inside building" -- "y2";
    "east pit" -- "east end of twopit room";
    "west end of twopit room" -- "west pit";
    "swiss cheese room" -- "east end of twopit room";
    "dirty passage" -- "dusty rock room";
    "dirty passage" -- "low n/s passage";
    "bedquilt" -- "complex junction";
    "swiss cheese room" -- "bedquilt";
}
        \end{lstlisting} \\
        \hline
        \textbf{Incorrect} \\
        \hline
        \begin{lstlisting}[basicstyle=\ttfamily\tiny]
digraph G {
    Plover Room -> Y2 [style=invis];
    Y2 -> Low N/S Passage [style=invis];
    Low N/S Passage -> Dirty Passage [style=invis];
    Dirty Passage -> Dusty Rock Room [style=invis];
    Dusty Rock Room -> Complex Junction [style=invis];
    Complex Junction -> Bedquilt [style=invis];
    Bedquilt -> Swiss Cheese Room [style=invis];
    Swiss Cheese Room -> Twopit Room [style=invis];
    Twopit Room -> West Pit [style=invis];
    West Pit -> Twopit Room [style=invis];
    Twopit Room -> East Pit [style=invis];
}
        \end{lstlisting} \\
        \hline
    \end{tabular}
    \caption{Example of correct and incorrect graph descriptions.}
    \label{table:graphviz_comparison}
\end{table}

    \item \textbf{Naming ambiguity:} Sometimes models confuse or slightly change location names, which is inevitable, but normalization and fuzzy matching help a lot. Normalization includes lowercase, removal of articles, and removal of prepositions at the beginning of names (again, we recommend examining the metrics calculation code for more details). Fuzzy matching threats two names equal if one of them is a substring of another. These efforts are especially important in Task 1, because in other tasks we provide the model with the list of proper location names as a part of input. To illustrate the impact of normalization, we provide the scores calculated with and without preprocessing for several models in Table \ref{table:strict_fuzzy_comparison}. 

\begin{table}[t]
\small
    \centering
    \begin{tabular}{|l|c|c|c|}
        \hline
        \textbf{Model} & \textbf{strict} & \textbf{fuzzy} & \textbf{diff (pp)} \\
        \hline
        gpt-3-5-turbo            & 30.6\%  & 41.5\%  & +10.9 \\
        gpt-4-turbo   & 53.1\%  & 67.4\%  & +14.3 \\
        gpt-4o        & 52.2\%  & 66.9\%  & +14.7 \\
        claude-3-opus & 50.7\%  & 64.5\%  & +13.8 \\
        llama3-8b                 & 41.1\%  & 55.8\%  & +14.7 \\
        mixtral-8x7b              & 12.6\%  & 17.0\%  & +~4.4  \\
        llama3-70b                & 46.5\%  & 61.7\%  & +15.2 \\
        mixtral-8x22b             & 45.8\%  & 60.2\%  & +14.4 \\
        \hline
    \end{tabular}
    \caption{Performance with fuzzy matching (0-shot).}
    \label{table:strict_fuzzy_comparison}
\end{table}

    \item \textbf{Location hallucinations:} Sometimes models invent locations that were not visited in the narrative but were mentioned as something seen in the distance. In other cases (especially with smaller models), the model is not always able to grasp the concept of location and builds a generic nesting graph of the entities mentioned (including containers and objects); see Appendix \ref{sec:appendixC} for an example. In both cases, this leads to additional false nodes in the graph, which can be illustrated by degradation precision while keeping recall on a relatively high level (see Table \ref{table:performance_comparison}).

\begin{table}[t]
\small
    \centering
    \begin{tabular}{|l|c|c|c|}
        \hline
        \textbf{Model} & \textbf{F1} & \textbf{recall} & \textbf{precision} \\
        \hline
        gpt-3-5-turbo            & 41.5\%  & 62.5\%  & 34.8\% \\
        gpt-4-turbo   & 67.4\%  & 72.3\%  & 65.7\% \\
        gpt-4o        & 66.9\%  & 75.2\%  & 62.5\% \\
        claude-3-opus & 64.5\%  & 73.9\%  & 59.6\% \\
        \hline
    \end{tabular}
    \caption{F1/recall/precision across models (0-shot).}
    \label{table:performance_comparison}
\end{table}

\end{itemize}

\section{Discussion}

The task of defining and identifying locations and paths within a narrative is inherently complex and somewhat subjective. This complexity raises several questions about the nature and definition of spatial objects and relations in fictional texts.

\begin{figure*}[t]
  \includegraphics[width=\textwidth]{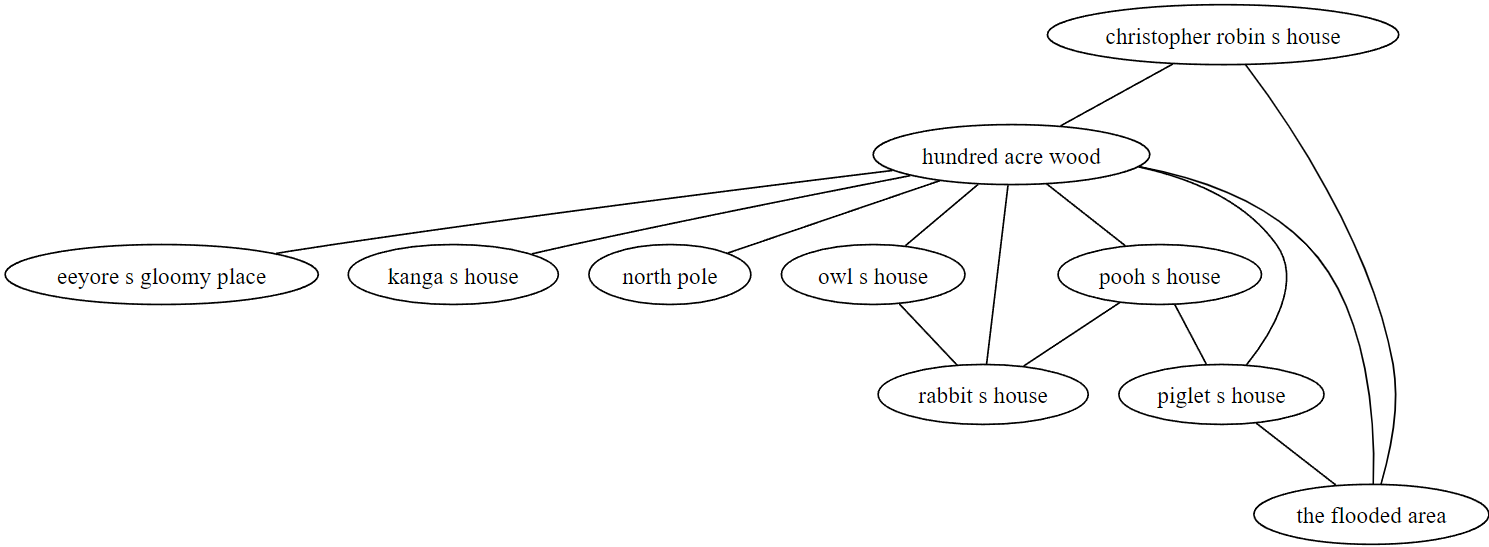}
  \caption{A spatial graph reconstructed from a text of the Winnie-the-Pooh book.}
  \label{fig:pooh-map}
\end{figure*}

\begin{figure}[t]
  \includegraphics[width=\columnwidth]{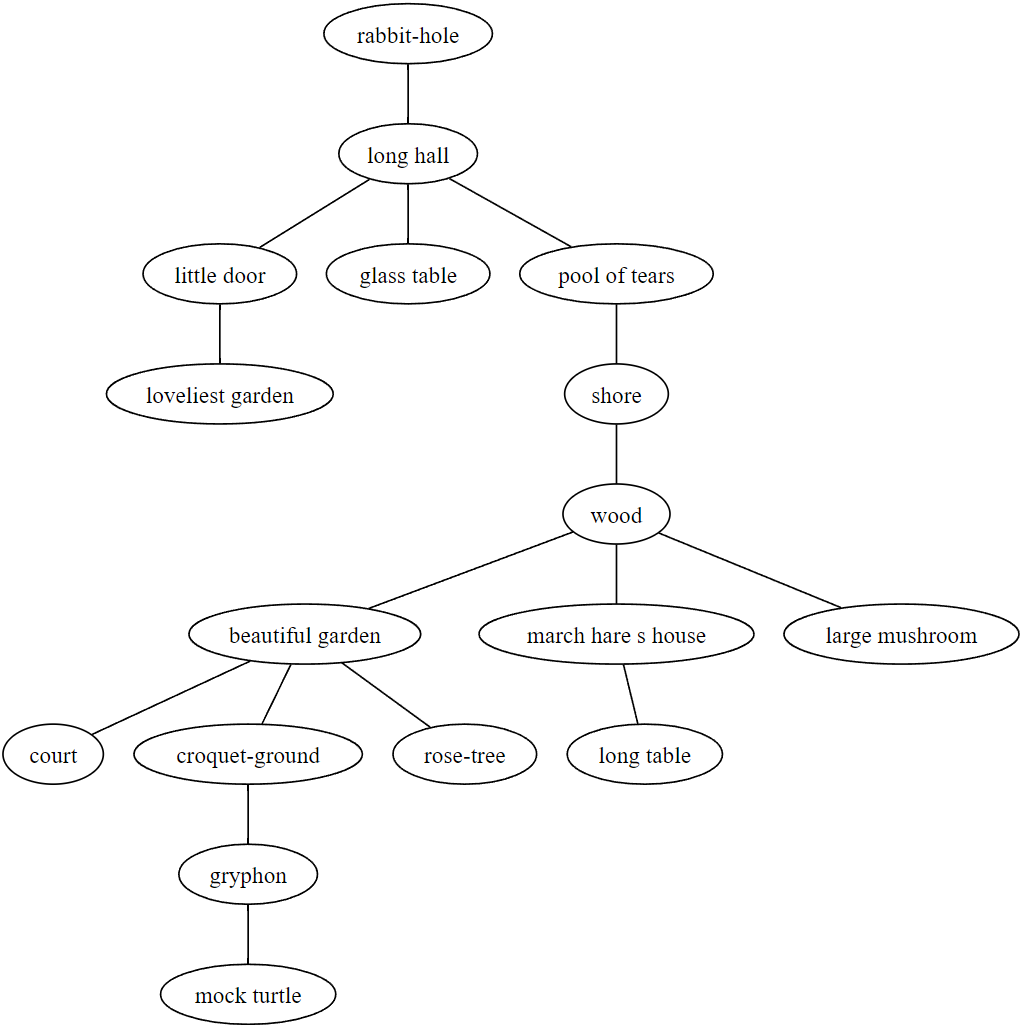}
  \caption{A spatial graph reconstructed from a text of the Alice in Wonderland book.}
  \label{fig:alice-map}
\end{figure}

Firstly, one might question whether locations can be nested. For instance, is a closet within a grandmother's room, where Little Red Riding Hood hid, a separate location or merely an object within a location? This ambiguity extends to whether an encompassing location should be described as a subgraph or just as an intermediate location connecting other ones; see, for example, the Hundred Acre Wood in Winnie-the-Pooh (Figure \ref{fig:pooh-map}). Another debatable aspect is the distinction between a location and a container. Can a place like a "glass table" be considered a location? This might depend on the size of the characters and their ability to move around, as seen in Alice in Wonderland (see Figure \ref{fig:alice-map}). Moreover, should a location be immobile? Seems like it's not necessary, as locations can include a hot air balloon basket, an elevator, or a cab transporting characters during their conversation. The concept of directly connected locations also requires discussion. If a character travels from home to work, how detailed should the description of their intermediate movements be?

Despite the lack of universal answers to these questions, an average reader can intuitively respond to them without much thought. Modern LLMs, as demonstrated, can also reproduce the author's intended graph with sufficient accuracy (F1-score up to 78+\%).

This observation leads us to the idea that, despite the apparent subjectivity and arbitrariness of definitions, our perception of space described in a narrative is governed by more formal and predictable principles. For example, the description reporting bias (the principle of omitting trivial things) suggests that writers tend to skip obvious information while writing texts. In \cite{Fludernik}, the author explores how narratives mimic natural conversation by selectively including or omitting details to focus on what is essential for the story. She argues that this selective reporting aligns with how people naturally communicate, highlighting significant events while excluding the mundane. Similarly, Chekhov's gun principle \cite{tikhonov2022actionable} implies that mentioned objects should be important to the plot.

Thus, while the task of defining and identifying locations in narratives is complex and subjective, our findings suggest that there are underlying principles guiding our perception of space in narratives. These principles can be leveraged to assess and improve the performance of LLMs in spatial reasoning tasks, as demonstrated by the results of our benchmark. Future work should continue to explore these principles and refine the methods for evaluating and enhancing the spatial reasoning capabilities of LLMs.

\section{Conclusion}

In this paper, we introduced PLUGH, a benchmark designed to evaluate the spatial reasoning capabilities of LLMs based on fiction texts. Our evaluation of various models reveals significant variability in performance across different tasks, highlighting the strengths and weaknesses of current LLMs in spatial reasoning. Despite recent advancements, there is substantial room for improvement, particularly in addressing common issues such as incorrect formatting, naming ambiguity, and location hallucinations.

Our findings underscore the importance of diverse benchmarks and robust error analysis techniques in advancing the field of spatial reasoning. Future research should focus on developing models with enhanced spatial reasoning capabilities and refining evaluation methodologies to provide a comprehensive assessment of model performance.

\section*{Acknowledgments}
We would like to thank the developers of the Jericho emulator and the authors of the text-based games used in this study.


\clearpage
\appendix

\section{Conversion Prompt}
\label{sec:appendixA}

The GPT-4 prompt we used to convert transcripts into fiction texts:

\begin{small}
\begin{verbatim}

{"role": "system", "content": "You are the extremely
powerful AI-powered fiction text writer. You will be
provided with a transcript of an interactive fiction
game - some player sends actions, and a game returns
observation. Your task is to rewrite it into a good-
looking fiction text."},
{"role": "user", "content": "# Transcript <...>"},
{"role": "system", "content": """Rewrite it into a 
good-looking fiction text -- put it into third 
person  PoV,   replace  artificial   looking  text 
structures with normal narrative, remove redundant 
and repeating pieces.
KEEP MENTIONS OF  LOCATION NAMES DURING TRAVELS."""},
\end{verbatim}
\end{small}
\section{Example of rewriting results}
\label{sec:appendixB}

An example of the text before and after rewriting.
\\ \\
Before:
\begin{small}
\begin{quote}
<...>

Tunnel

You are in a narrow tunnel whose entrance has been blocked by gigantic, translucent boulders. Your only option is to continue to the northeast, where the tunnel gradually widens.

> examine boulders

The boulders are, in reality, milky white crystals and fairly regular in shape, like giant diamonds.

> ne

Cultural Complex

This imposing ante-room, the center of what was apparently the cultural center of the GUE, is adorned in the ghastly style of the GUE\'s "Grotesque Period."  With leering gargoyles, cartoonish friezes depicting long-forgotten scenes of GUE history, and primitive statuary of pointy-headed personages unknown (perhaps very, very distant progenitors of the Flatheads), the place would have been best left undiscovered. North of here, a large hallway passes under the roughly hewn inscription "Convention Center."  To the east, under a fifty-story triumphal arch, a passageway the size of a large city boulevard opens into the Royal Theater. A relatively small and unobtrusive sign (perhaps ten feet high) stands nearby. South, a smaller and more dignified (i.e. post-Dimwit) path leads into what is billed as the "Hall of Science."

> n

Convention Center Lobby

You are standing at the entrance of the GUE Convention Center. <...>
\end{quote}
\end{small}
After:
\begin{small}
\begin{quote}
<...> giant, translucent boulders tumbled from the ceiling, effectively sealing them inside. The only path forward lay to the northeast, where the tunnel gradually widened, beckoning them deeper into the unknown.

The boulders, upon closer inspection, revealed themselves as milky white crystals, shaped like enormous, uncut diamonds. Pushing forward, the adventurer entered the Cultural Complex, a grandiose ante-room adorned in the grotesque style of the GUE's 'Grotesque Period'. Leering gargoyles and cartoonish friezes depicted obscure scenes of forgotten history, surrounding primitive statues of pointy-headed figures. To the north, a hallway led to the Convention Center; to the east, a grand passageway opened into the Royal Theater.

Venturing north, the adventurer found themselves in the lobby of the Convention Center. <...>
\end{quote}
\end{small}

\section{All non-isomorphic graphs}
\label{sec:appendixD}

\begin{figure}[h]
  \includegraphics[width=\columnwidth]{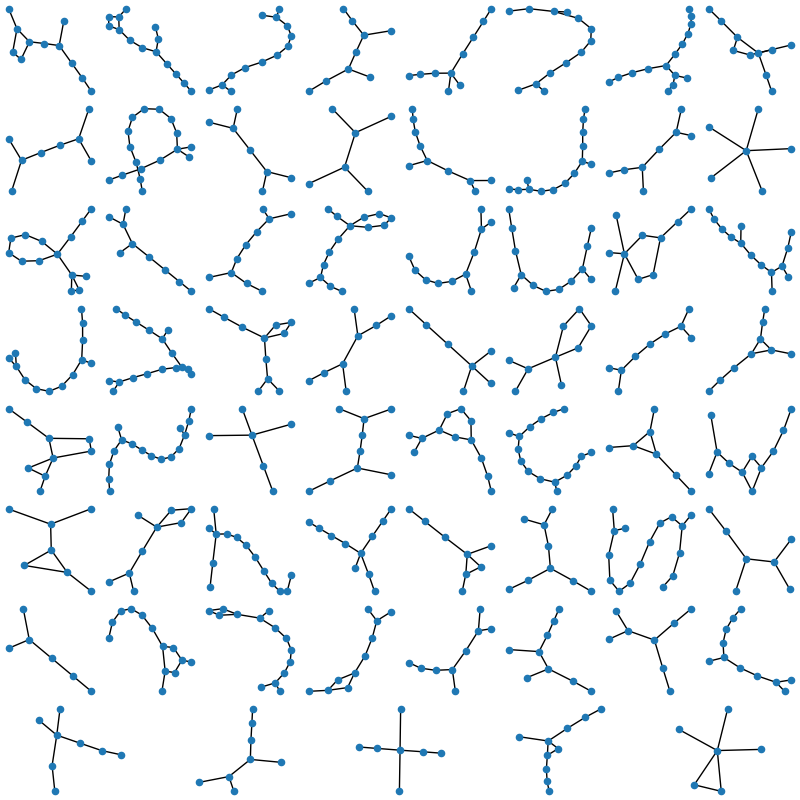}
  \caption{61 mutually non-isomorphic graphs used in our benchmark.}
  \label{fig:non-isomorphic}
\end{figure}

\newpage
\onecolumn
\section{Hallucinated locations example}
\label{sec:appendixC}

\begin{figure}[h]
  \includegraphics[width=.58\columnwidth]{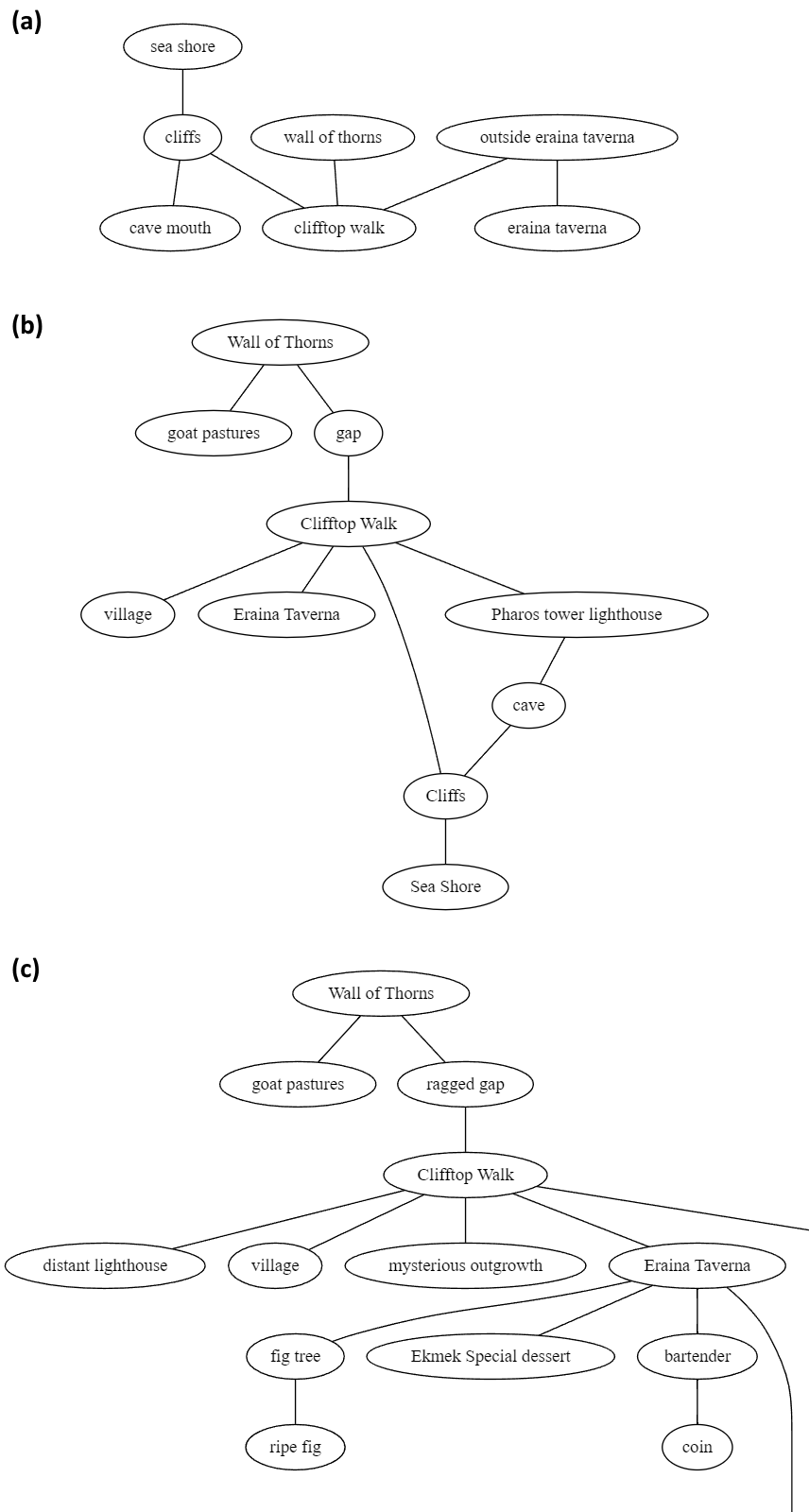}
  \caption{Several graphs for a snippet from the Curses game:\\
  (a) ground truth graph; \\
  (b) graph with extra hallucinated locations; \\
  (c) graph with objects as extra nodes (truncated).}
  \label{fig:hl}
\end{figure}

\end{document}